\documentclass[10pt,twocolumn,letterpaper]{article}
\pdfoutput=1
\usepackage{iccv}
\usepackage{times}
\usepackage{epsfig}
\usepackage{graphicx}
\usepackage{amsmath}
\usepackage{amssymb}

\usepackage{bm}
\usepackage{capt-of}
\usepackage{mathtools}
\usepackage{siunitx}
\usepackage[figurename=Figure]{caption}
\usepackage{scrextend}
\usepackage{xspace}
\usepackage{xcolor}
\usepackage{multirow}
\usepackage{wrapfig}
\usepackage{amsthm}
\usepackage{thmtools}
\usepackage{algorithm,algpseudocode}
\usepackage[export]{adjustbox}
\usepackage{url}
\usepackage{lineno}
\usepackage{pifont}
\usepackage{booktabs}

\newcounter{algsubstate}

\DeclareMathOperator*{\argmin}{arg\,min}

\newtheorem*{theorem*}{Theorem}

\renewcommand{\eqref}[1]{\mbox{Eq.~(\ref{#1})}}

\makeatletter
\DeclareRobustCommand\onedot{\futurelet\@let@token\@onedot}
\def\@onedot{\ifx\@let@token.\else.\null\fi\xspace}
\def\eg{\emph{e.g}\onedot}
\def\ie{\emph{i.e}\onedot}

\makeatother

\newcolumntype{L}[1]{>{\raggedright\arraybackslash}p{#1}}
\newcolumntype{C}[1]{>{\centering\arraybackslash}p{#1}}
\newcolumntype{R}[1]{>{\raggedleft\arraybackslash}p{#1}}

\newcommand{\cmark}{\ding{51}}%
\newcommand{\xmark}{\ding{55}}%
\newcommand*\samethanks[1][\value{footnote}]{\footnotemark[#1]}

\usepackage[pagebackref=true,breaklinks=true,letterpaper=true,colorlinks,bookmarks=false]{hyperref}

\iccvfinalcopy 

\def\iccvPaperID{4568} 

\ificcvfinal\fi

\begin{document}

\title{Harvard Glaucoma Detection and Progression: A Multimodal Multitask Dataset and Generalization-Reinforced Semi-Supervised Learning}

\author{Yan Luo\thanks{First three authors contribute equally.} \quad Min Shi\samethanks \quad Yu Tian\samethanks \quad Tobias Elze \ \  Mengyu Wang \\
Harvard Ophthalmology AI Lab, Harvard University\\
{\tt\small \{yluo16, mshi6, ytian11, tobias\_elze, mengyu\_wang\}@meei.harvard.edu}
}


\maketitle
\ificcvfinal\thispagestyle{empty}\fi

	
	
	
\begin{abstract}
Glaucoma is the number one cause of irreversible blindness globally.
A major challenge for accurate glaucoma detection and progression forecasting is the bottleneck of limited labeled patients with the state-of-the-art (SOTA) 3D retinal imaging data of optical coherence tomography (OCT). To address the data scarcity issue, this paper proposes two solutions. First, we develop a novel generalization-reinforced semi-supervised learning (SSL) model called pseudo supervisor to optimally utilize unlabeled data.
Compared with SOTA models, the proposed pseudo supervisor optimizes the policy of predicting pseudo labels with unlabeled samples to improve empirical generalization.
Our pseudo supervisor model is 
evaluated with two clinical tasks consisting of glaucoma detection and progression forecasting. The progression forecasting task is evaluated both unimodally and multimodally. Our pseudo supervisor model demonstrates superior performance than SOTA SSL comparison models. Moreover, our model also achieves the best results on the publicly available LAG fundus dataset.  Second, we introduce the Harvard Glaucoma Detection and Progression (Harvard-GDP) Dataset, a multimodal multitask dataset that includes data from 1,000 patients with OCT imaging data, as well as labels for glaucoma detection and progression. This is the largest glaucoma detection dataset with 3D OCT imaging data and the first glaucoma progression forecasting dataset that is publicly available. Detailed sex and racial analysis are provided, which can be used by interested researchers for fairness learning studies. Our released dataset is benchmarked with several SOTA supervised CNN and transformer deep learning models. The dataset and code are made publicly available via 
\url{https://ophai.hms.harvard.edu/datasets/harvard-gdp1000}.
\vspace{-10pt}

\end{abstract}

\section{Introduction}

\label{intro}


Glaucoma is the leading cause of irreversible blindness globally caused by retinal nerve fiber layer (RNFL) damage\cite{tham2014global, quigley1996number, quigley2006number, sun2022time}. The global prevalence of glaucoma for the population between 40 and 80 years old is 3.54\% \cite{tham2014global}. Timely clinical treatment in the early stage of glaucoma can significantly reduce the risk of further vision loss \cite{conlon2017glaucoma, coleman2012advances}. However, most commonly, glaucoma patients are not aware of their early VF loss until the VF loss becomes severe enough to impair their daily activities such as reading and driving due to the brain and fellow eye compensation\cite{artes2005longitudinal, crabb2013does, kellman1991theory, morgan1982mechanisms, nelson2000predicting, hu2015patterns}.

%

\begin{figure}
  \centering
    \includegraphics[width=0.4\textwidth]{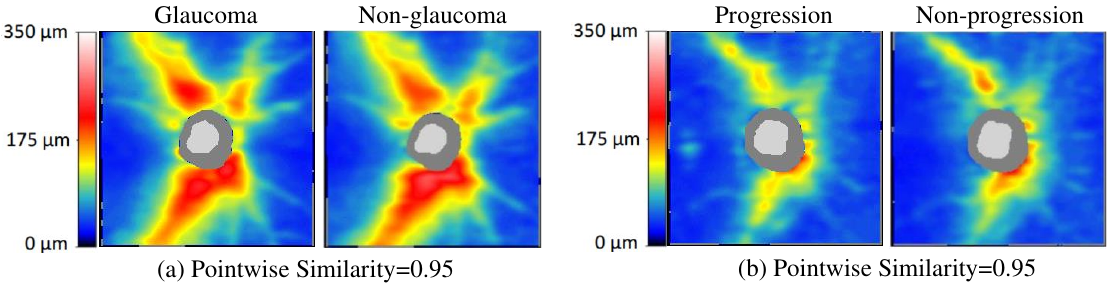}
     \vspace{-0.3cm}
  \caption{Examples of RNFLT maps with high similarities but from different label groups.} 
  \label{varsample}
\end{figure}

It is desirable to develop automated glaucoma screen tools to reduce societal disease burden \cite{varma2011assessment}. The clinical diagnosis of glaucoma is a holistic decision primarily based on RNFL damage and VF loss assessment supplemented with other patient clinical information and family disease history \cite{hood2007framework, malik2012structure}. The VF test is only available in the ophthalmology specialty. So existing automated screening tools are based on the idea of predicting glaucoma diagnosis from evaluating RNFL damage using retinal images \cite{chen2015automatic, phene2019deep, liu2019development, asaoka2019using, mehta2021automated, li2019attention, christopher2018performance, kihara2022policy, ran2019detection, wu2021performances}. Before the massive clinical adoption of the 3D retinal imaging tool of optical coherence tomography (OCT) a decade ago \cite{huang1991optical, bussel2014oct}, 2D fundus photos were used to assess RNFL damage. Therefore, most prior glaucoma detection models are based on the 2D fundus photo \cite{li2019attention, phene2019deep, keel2019visualizing, chen2015automatic, liu2019development, ahn2018deep, shibata2018development, christopher2018performance}, which is not the state of the art (SOTA) clinical imaging modality anymore. OCT has replaced fundus photos as the primary imaging tool for glaucoma clinical care due to its superiority in measuring RNFL damage \cite{medeiros2019machine, bussel2014oct}. There have been a number of recent works using OCT scans to automatically screen glaucoma with machine learning \cite{christopher2018performance, ran2019detection, asaoka2019using, kihara2022policy, fang2022dataset, wu2022gamma, mehta2021automated}. Apart from glaucoma detection, a more clinically significant task is progression forecasting using OCT scans \cite{nouri2021prediction, christopher2018retinal, kamalipour2022combining, herbert2022forecasting}, which could inform clinicians if they should treat the patient conservatively with eye drops or aggressively with invasive surgeries.

For both the glaucoma screening and progression forecasting tasks, a significant challenge is limited labeled patients with the SOTA OCT measurements as the fundus photo is already outdated in clinical practice \cite{christopher2018performance, ran2019detection, asaoka2019using, kihara2022policy, fang2022dataset, wu2022gamma, mehta2021automated}. This problem with the lack of labeled patients is even more severe for progression forecasting, which requires longitudinal VF measurements to determine progression outcomes \cite{nouri2021prediction, christopher2018retinal, kamalipour2022combining}. Therefore, there is a strong need for developing novel semi-supervised learning models to maximally utilize unlabeled patient data in clinics to improve glaucoma detection and progression forecasting. However, apart from several papers using standard semi-supervised learning methods for glaucoma detection \cite{diaz2019retinal, al2019semi, fan2020one, wang2020towards, zhang2010origa, bechar2018semi}, there has been no dedicated algorithm development for semi-supervised learning with its application in glaucoma. SOTA semi-supervised learning (SSL) approaches in computer vision are generally based on consistency learning \cite{berthelot2019mixmatch,sohn2020fixmatch,berthelot2019remixmatch,tarvainen2017mean} or pseudo-labeling \cite{liu2022acpl,rizve2020defense}. Unlike consistency learning, the pseudo-labeling approaches do not need domain knowledge to design the image augmentation strategy. While many prior works have endeavored to optimally select pseudo-labeled samples to improve the classifier training, the tasks of glaucoma detection and progression forecasting still have some unique challenges that might not be well tackled by existing pseudo-labeling SSL methods \cite{rizve2020defense,shi2018transductive,cascante2020curriculum,liu2022acpl, pham2021meta}.
Figure \ref{varsample} shows examples of RNFLT maps with high similarities but from different label groups. With such obscured group differences, existing pseudo-labeling SSL methods might not be able to provide sufficient guidance to select correct pseudo-labeled samples to improve the classifier training. Furthermore, having high-quality public datasets with OCT scans for glaucoma detection and progression forecasting is equally important to enable more computer vision researchers to study this topic to push forward the field,


In this study, we make two contributions: (1) developing a generalization-reinforced semi-supervised learning model called pseudo supervisor to improve glaucoma detection and progression forecasting. Specifically, our pseudo supervisor model will select pseudo-labeled samples via a policy net that optimizes generalization error with a validation subset to update the classifier. (2) releasing a pilot multitask and multimodal glaucoma dataset, called Harvard Glaucoma Detection and Progression (Harvard-GDP) Dataset, for computer vision researchers to study this topic. In Table\ref{tbl:problem}, we have listed major public glaucoma datasets. Most of them are fundus photo datasets, while there are two major OCT datasets previously published \cite{fang2022dataset,wu2022gamma}. In comparison, our dataset has the \textit{largest} patient numbers among all datasets with patient numbers available. In addition, our dataset is the \textit{first} and the only one with progression forecasting task data, which is a way more clinically significant task than glaucoma detection itself. Compared with the other two OCT datasets, our dataset is superior in patient numbers (1,000 versus up to 300), task versatility (glaucoma detection and progression forecasting versus glaucoma detection only), and availableness of objective visual function tests (i.e. VF test). Furthermore, since our dataset is from the US population, the racial representation in our data is more diverse than the two OCT datasets \cite{fang2022dataset,wu2022gamma} and the large LAG dataset \cite{Li_CVPR_2019}, which consists of mainly Asian patients. Therefore, our dataset can be potentially used for medical fairness learning studies, especially since we know Black people have more than doubled glaucoma prevalence than other races, which is a significant disparity \cite{rudnicka2006variations, friedman2006prevalence}.

\begin{table}
	\centering
	\caption{\label{tbl:problem}
	   Public Glaucoma Datasets.
	}
  \vspace{-0.3cm}
	\adjustbox{width=1\columnwidth}{
	\begin{tabular}{C{15ex} L{10ex} L{12ex} C{10ex} C{12ex} C{10ex} C{20ex} C{16ex}}
		\toprule
		\textbf{Study} & \textbf{Imaging Modality} & \textbf{Sample Size} & \textbf{Glaucoma Detection} & \textbf{Progression Forecasting}  & \textbf{Multimodal} & \textbf{Label Source}&\textbf{Accessibility}\\
		\cmidrule(lr){1-1} \cmidrule(lr){2-2} \cmidrule(lr){3-3} \cmidrule(lr){4-4}\cmidrule(lr){5-5}\cmidrule(lr){6-6}\cmidrule(lr){7-7}\cmidrule(lr){8-8}
	
            LAG \cite{Li_CVPR_2019} & Fundus & NA (4,854) & \cmark & \xmark  & \xmark & Clinician Assessment & Upon Request\\
            REFUGE \cite{orlando2020refuge} & Fundus & NA (1,200) & \cmark &\xmark  & \xmark & Clinician Assessment &Web Download\\
            G1020 \cite{bajwa2020g1020} & Fundus & 432 (1,020) & \cmark & \xmark & \xmark & Clinician Assessment & Web Download\\
            ACRIMA \cite{diaz2019cnns} & Fundus & NA (705) & \cmark & \xmark & \xmark & Clinician Assessment & Web Download\\
             RIM-ONE \cite{batista2020rim} & Fundus & 169 (485) & \cmark & \xmark & \xmark & Clinician Assessment & Web Download\\
             ORIGA \cite{zhang2010origa} & Fundus & NA (650) & \cmark & \xmark & \xmark & Clinician Assessment & Upon Request\\
             PAPILA \cite{kovalyk2022papila} & Fundus & 244 (488)  & \cmark & \xmark & \xmark & Clinician Assessment & Web Download\\
              GOALS \cite{fang2022dataset} & OCT & 158 (300) & \cmark & \xmark & \xmark & Clinician Assessment & Web Download\\
               GAMMA \cite{wu2022gamma} & OCT & 300 (300) & \cmark & \xmark & \cmark & Clinician Assessment & Web Download\\

          \midrule
    Ours & OCT & 1,000 (1,000) & \cmark & \cmark & \cmark & Functional Test & Web Download\\
		\bottomrule	
	\end{tabular}}
\end{table}

    

\section{Related Work}

\noindent \textbf{Glaucoma Detection}. Vast majority of glaucoma detection works using deep learning are based on the fundus photo, which is a 2D snapshot of the patient's retina \cite{li2019attention, phene2019deep, keel2019visualizing, chen2015automatic, liu2019development, ahn2018deep, shibata2018development, christopher2018performance,tian2021constrained,tian2021self}. However, the 3D imaging modality OCT scan has replaced fundus photos to become the de facto standard for structural damage assessment in glaucoma. Compared with the deep learning glaucoma detection using fundus photos, deep learning glaucoma detection works using OCT scans generally have much smaller sample sizes, which are up to several thousand \cite{christopher2018performance, ran2019detection, asaoka2019using, kihara2022policy, fang2022dataset, wu2022gamma, mehta2021automated}. As deep learning models are typically data-hungry, there is a clear need to develop new approaches to deal with the limited labeled data issue. It is known that semi-supervised learning can improve prediction accuracy by leveraging unlabeled data in model training. So far, apart from several glaucoma detection papers using standard semi-supervised learning models \cite{diaz2019retinal, al2019semi, fan2020one, wang2020towards, zhang2010origa, bechar2018semi}, no novel semi-supervised learning model has been developed with an application focus on glaucoma.

\noindent \textbf{Progression Forecasting}. Compared with the glaucoma detection task, progression forecasting is a much more clinically significant task for patients. If we could have an accurate progression forecasting model to inform clinicians of which patient will progress or not, then the clinicians can accordingly decide if a patient should be treated more aggressively with invasive surgeries, which may have a strong side effect, or more conservatively with eye drops. So far, due to the data scarcity issue, we have only found several recent papers using retinal imaging data (whether fundus or OCT) to forecast progression \cite{nouri2021prediction, christopher2018retinal, kamalipour2022combining}. The sample sizes of these works are only up to several hundred. Imaginably, we have not found any prior studies applying semi-supervised learning for progression forecasting.

\noindent \textbf{Semi-Supervised Learning}.
Semi-Supervised Learning (SSL) aims to optimize a model with both labeled and unlabeled data. The current SSL methods can be divided into pseudo-labeling-based methods \cite{rizve2020defense,shi2018transductive,cascante2020curriculum,liu2022acpl} and/or consistency-based methods \cite{tarvainen2017mean,sohn2020fixmatch,berthelot2019mixmatch,berthelot2019remixmatch,liu2021self,liu2022perturbed}. Consistency-based SSL methods optimize the model with standard classification losses (e.g., cross-entropy) on labeled images and regularizes/minimizes the prediction outputs of different weakly and/or strongly augmented views of unlabeled images, where these views are constructed using different types of image augmentations, such as flip-and-shift, color sharpness, and contrast, etc. Even though consistency-based SSL methods achieve SOTA results in many computer vision benchmarks \cite{tarvainen2017mean,sohn2020fixmatch,berthelot2019mixmatch,berthelot2019remixmatch}, these methods significantly rely on the orderly design of augmentation functions that requires domain knowledge to design proper image augmentation strategies, which are even more challenging for multimodal modeling.
Prior pseudo-labeling SSL methods \cite{rizve2020defense,shi2018transductive,cascante2020curriculum,liu2022acpl, pham2021meta} typically utilize a model that trained with the labeled subset to estimate the pseudo labels of the confident unlabeled samples and take these pseudo-labeled data to re-train the model. As mentioned previously, pseudo-labeling-based SSL approaches is a general SSL learning framework for different data modalities naturally. More importantly, as mentioned in Section \ref{intro}, since the structural differences between glaucoma and non-glaucoma (similarly for progression versus non-progression) are obscured as shown in Figure \ref{varsample}, existing pseudo-labeling SSL
methods \cite{rizve2020defense,shi2018transductive,cascante2020curriculum,liu2022acpl, pham2021meta} might not be able to provide sufficient guidance to select correct pseudo labeled samples to benefit the classifier training. In this aspect, we propose a generalization-reinforced SSL model by selecting optimal pseudo-labeled samples for classifier training that explicitly optimize generalization error with a policy net stochastically generates pseudo labels.



\noindent \textbf{Reinforcement Learning}.
Reinforcement learning studies an agent interacts with the environment and learns an optimal policy to tackle decision-making problems effectively \cite{Sutton_NeurIPS_1999,Baxter_JAIR_2001,Ghavamzadeh_CSDFPS_2003,Silver_ICML_2014,Silver_Nature_2017}. Reinforcement learning has been applied to semi-supervised learning setting for real-world applications, \eg domain adaptation \cite{Liu_BMVC_2020}. However, how to utilize reinforcement learning techniques to determine pseudo labels for improving the generalization of classification models remains under exploration.

\section{Multimodal and Multitask Dataset}
\label{data_section}

\subsection{Data Collection and Quality Control}
This work was approved by the institutional review board (IRB) and adhered to the tenets of the Declaration of Helsinki. Because of the retrospective nature of the study, the IRB waived the need for informed consent of patients. 
The glaucoma patients are from Mass Eye and Ear of Harvard Medical School, which were tested between 2010 and 2021. The RNFLT ($0-350$ microns) is from the OCT report provided by the OCT device manufacturer (Cirrus, Carl Zeiss Meditec, Jena, Germany) calculated as the vertical distance between the internal limiting membrane boundary and nerve fiber layer boundary. The RNFLT measurement is the de facto structural measurement for clinicians to diagnose glaucoma. Each two-dimensional $224\times 224$ RNFLT map is generated within the $6 \times 6$ mm$^2$ peripapillary area around the optic nerve head. The VF testing is performed using the Humphrey Field Analyzer with the 24-2 protocol, which measures the patient's eye's visual spatial sensitivity with a radius of 24 degrees from the fixation. Data quality control based on standard clinical practice guideline following the manufacturer's recommendation was applied to exclude unreliable data: (1) OCT scans with signal strength less than 6 (10 represents the best imaging quality) were excluded; (2) VF tests with fixation losses $>$ 33\%, false positive rates $>$ 20\% or false negative rates $>$ 20\%  were excluded. Each VF measurement is represented as a vector of 52 total deviation (TD) values between -38 dB and 26 dB. In this work, we refer RNFLT map and VF test as two data modalities which constitute a multimodal dataset. Note that, the modifier word ``Harvard" from the Harvard GDP Dataset only indicates that our dataset is from the Department of Ophthalmology of Harvard Medical School and does not imply an endorsement, sponsorship, or assumption of responsibility by either Harvard University or Harvard Medical School as a legal identity.


\begin{figure}
  \centering
    \includegraphics[width=0.486\textwidth]{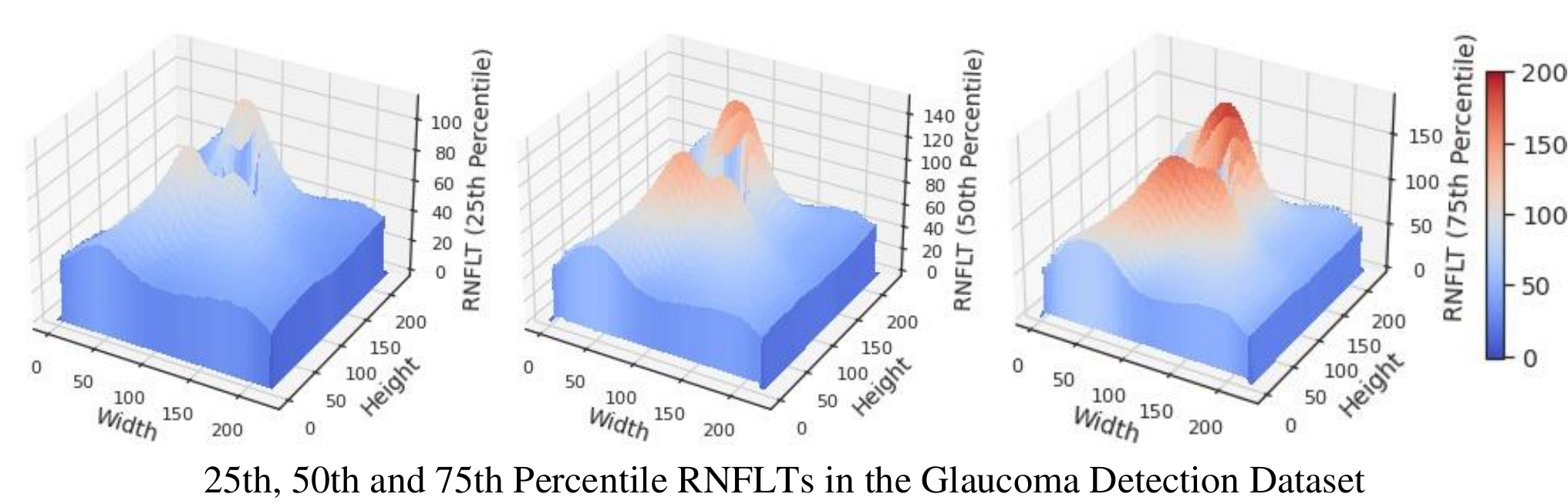}
    \vspace{-0.5cm}
  \caption{Pointwise RNFLT distribution at different percentiles.} 
   \vspace{-0.7cm}
  \label{rnfltdist}
\end{figure}


\subsection{Data Analysis}

The glaucoma detection dataset contains 15,725 training samples, 1,047 validation samples, and 4,287 samples from 5,640, 376, and 1,505  glaucoma patients, respectively. Half of the samples are unlabeled. Among the half samples with labels, 45.2\% of the samples are glaucoma, while the rest are non-glaucoma. The progression forecasting dataset includes 5,889 training samples, 431 validation samples, and 1,547 test samples from 1,292, 86, and 345 glaucoma patients, respectively. Among the half samples with progression outcome labels, 24.0\% and 2.6\% of the samples are TD Progression and MD Fast Progression, respectively. The two clinically accepted criteria to define glaucoma progression: (1) TD Progression: eyes with at least three locations with TD slope $\leq$ -1 dB; (2) MD Fast Progression: eyes with MD slope $\leq$ -1 dB \cite{vesti2003comparison}. The MD Fast Progression is more detrimental for the patients than TD progression \cite{anderson2019comparison}.  

Figure \ref{rnfltdist} presents the pointwise distributions of 25th, 50th, and 75th percentile RNFLTs in the train (validation set included) and test sets, which demonstrate two obvious superotemporal and inferotemporal RNFL bundles with their damages are known closely related to the visual function loss. 
As in the two representative examples shown in Figure \ref{varsample}, RNFLT maps with similar appearances may have different labels. This suggests the class boundary may be obscure to determine, which naturally prompts us to adopt semi-supervised solutions to address the glaucoma detection and progression forecasting problems which fully leverage the abundant information in unlabeled samples to mitigate the challenges.



The released dataset including 1,000 patients is a random subset from the entire dataset used to build the glaucoma detection and progression forecasting models in this paper, it naturally inherits the original data characteristics. The released glaucoma detection dataset contains 1,000 samples from 1,000 patients, where each sample includes two data modalities (i.e. RNFLT map and VF test) and respective glaucoma labels. The progression forecasting dataset is a subset from the released glaucoma detection dataset, which entails 500 patients with progression labels and patient demographics including age, sex, and race. For all glaucoma patients, the age distribution of 25th, 50th, and 75th percentiles are 54.3, 64.5, and 72.8 years, respectively. 
The male and female account for 56\% and 44\% of the 1,000 patients, respectively. Among the 1,000 patients, the White, Black, and Asian groups account for 76.4\%, 14.9\%, and 8.7\%, respectively. To the best of our knowledge, this is the first public glaucoma dataset with the patient demographic information available, which carries great potential to enable interested researchers to conduct and evaluate fairness learning studies.


\begin{figure}[!t]
	\centering
	\includegraphics[width=0.5\textwidth]{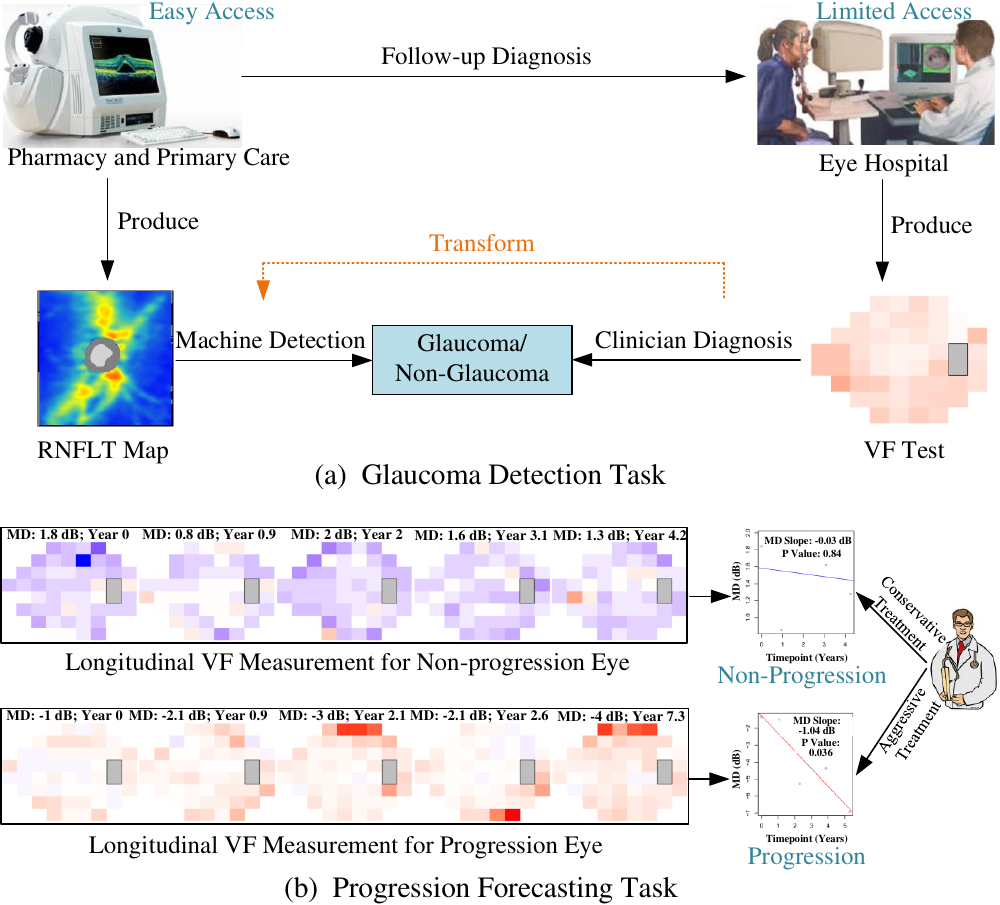}
  \vspace{-0.7cm}
	\caption{\label{fig:task}
    	The definitions of (a) glaucoma detection task; and (b) progression forecasting task. In the VF test, blue indicates normal visual function while red indicates abnormal visual function.
    	}
\end{figure}

\section{Problem Formalization \& Background}
\label{sec:problem}

In this section, we formally introduce the problems of glaucoma detection and progression forecasting. 

Denote $\mathcal{D}^{l}=\{(x_{i}, y_{i})| 1 \le i \le N^{l} \}$ be a labeled dataset, where $x\in \mathrm{R}^{n}$ are the RNFLT maps and $y\in \mathcal{Y}$ are corresponding glaucoma/progression labels. Also, we have a stack of unlabeled RNFLT maps, \ie $\mathcal{D}^{ul}=\{x_{j}|1 \le j \le N^{ul} \}$. We assume that we have a classification model $f: \mathrm{R}^{n} \xrightarrow{\theta} \mathcal{Y}$ ($f_{\theta}$ for short), where $\theta$ are the parameters. To update the classification model with the labeled data, a pre-defined loss function $\ell: \mathcal{Y} \xrightarrow{} \mathcal{Y}$ is used to gauge the discrepancy between the predictions and the ground-truth labels such that the classification model can be updated along the direction of minimizing the loss, \ie $min_{\theta} \ell (f_{\theta}(x), y)$.

As shown in Figure \ref{fig:task} (a), the practical utility scenario of glaucoma detection with deep learning is to use OCT scans that can be easily measured in the setting of primary care or pharmacy to predict glaucoma diagnosis defined by visual field tests that typically only can be measured in the setting of an ophthalmology clinic or eye hospital with specialized technicians. The clinical relevance of progression forecasting as shown in Figure \ref{fig:task} (b) with deep learning is that clinicians can better decide if a patient should be treated more conservatively with eye drops or aggressively with invasive surgeries. As suggested by conventional semi-supervised learning methods \cite{diaz2019retinal,al2019semi,liu2022acpl,sohn2020fixmatch,berthelot2019mixmatch}, pseudo labels $\tilde{y}$ are predicted by a supervisor model $\pi: \mathbf{R}^{n} \xrightarrow{\omega} \mathcal{Y}$ with the unlabeled data $\tilde{x}$, \ie $\tilde{y} = \pi(\tilde{x};\omega)$. Similar to the supervised learning setting, the unlabeled data with the pseudo labels can participate in the supervised learning process, namely $\min_{\theta} \ell (f_{\theta}(\tilde{x}), \tilde{y})$.



\begin{figure}
	\centering
	\includegraphics[width=.45\textwidth]{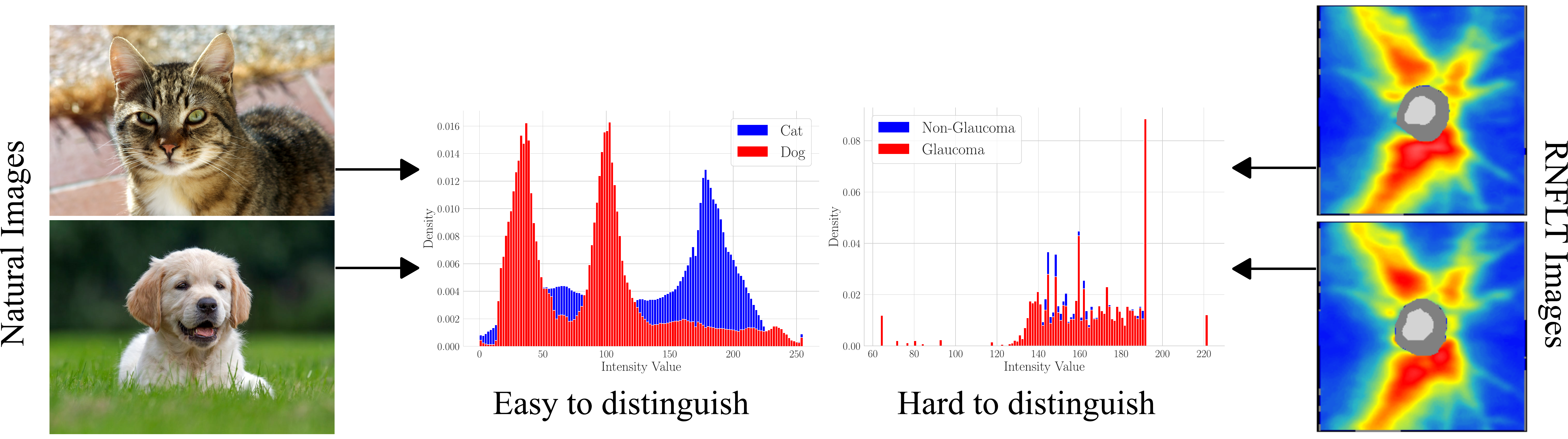}
  \vspace{-0.3cm}
	\caption{\label{fig:distinguish}
    	A challenge in glaucoma detection and progression forecasting. Unlike the classification with natural images, where the images with different labels are likely visually distinguishable, RNFLT images with different labels (\eg glaucoma vs. non-glaucoma and progression vs non-progression) might not be easily distinguishable.
    	}
      \vspace{-0.3cm}
\end{figure}

However, as shown in Figure \ref{fig:distinguish}, the natural images associated with different visual concepts are relatively easy to be distinguished. In contrast, the differences in RNFLT maps between glaucoma versus non-glaucoma (similarly for progression versus non-progression) can be very subtle due to large retinal physiological inter-subject variations. There might not be clear and adequate guidance in the learning process to determine the underlying labels of the unlabeled RNFLT maps. Instead, a more sensible way to better determine the pseudo labels is to apply a policy to obtain pseudo labels. Then, we observe how these pseudo labels affect the training process of the classifier and adjust the policy to predict the pseudo labels such that the generalizability of the classifier learned with the pseudo labels can be maximized. In short, the pseudo labels predicted by the supervisor model should be aligned with the improvement of the generalizability of the classifier on a set of unseen data $\mathcal{D}^{unseen}$. This objective can be formulated as
\begin{equation}
    \begin{split}
    & \min_{\{\tilde{y}\}} \frac{1}{|\mathcal{D}^{unseen}|} \sum_{(x,y)\in \mathcal{D}^{unseen}} \ell(f_{\theta^{*}}(x), y) \\
    &\text{where } \theta^{*} \in \argmin_{\theta} \frac{1}{|\mathcal{D}^{l+ul}|} \sum_{(x,y)\in \mathcal{D}^{l+ul}} \ell(f_{\theta}(x), y), 
\end{split}
\end{equation}
$\mathcal{D}^{l+ul}$ is the union of labeled dataset $\mathcal{D}$ and unlabeled dataset with pseudo labels $\mathcal{D}^{ul}$.




\begin{figure*}[!t]
	\centering
	\includegraphics[width=0.75\textwidth]{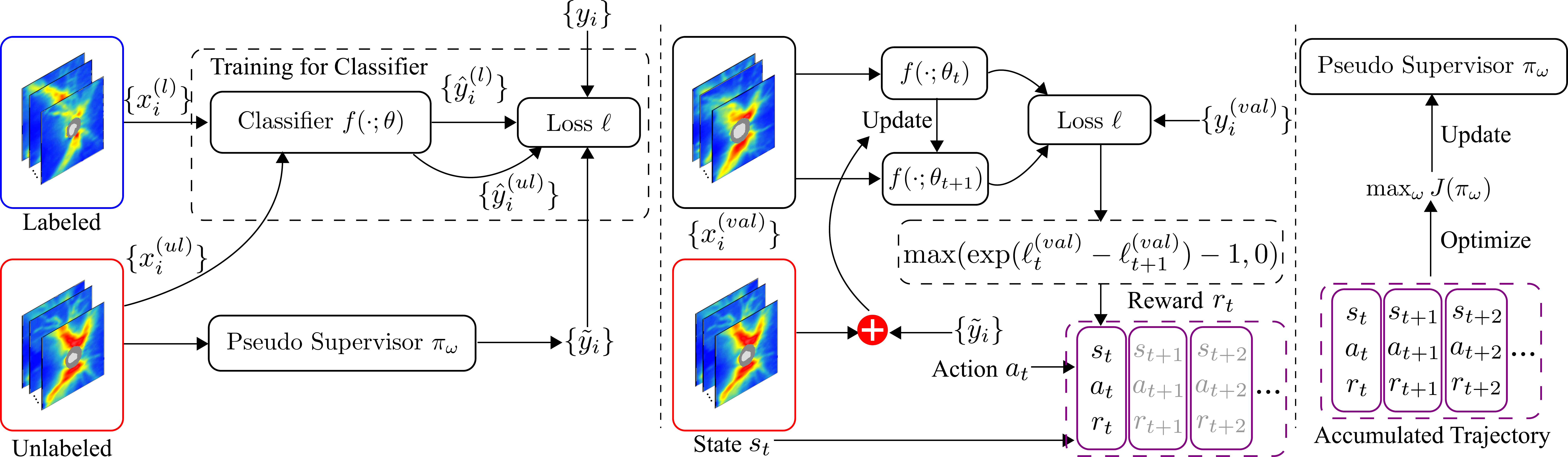}
 \vspace{-0.3cm}
	\caption{\label{fig:overview}
    	An overview of the proposed generalization-reinforced pseudo supervisor in the semi-supervised learning framework. The learning framework consists of the process of training the classifier, the process of measuring the empirical generalizability of the classifier to be used as rewards, and the process of updating the proposed pseudo supervisor to predict pseudo labels that can maximize the classifier's empirical generalizability.
    	}
\end{figure*}

\section{Generalization-Reinforced Pseudo Supervisor}
\label{sec:method}

Different from prior-work semi-supervised learning methods \cite{sohn2020fixmatch,berthelot2019mixmatch,berthelot2019remixmatch,oh2021distribution}, the proposed method explicitly leverages the empirical generalizability of the classifier that is quantified with unseen RNFLT maps (\eg the ones from the validation set) to optimize the policy of predicting the pseudo labels. The empirical generalizability can be viewed as a type of informative reward as the goal of training a classifier is to find one that can generalize well on the unseen test data drawn from the underlying distribution.

Figure \ref{fig:overview} shows how the proposed generalization-reinforced pseudo supervisor predicts pseudo labels and is trained according to the observed sets of states (maps), actions (pseudo labels), and rewards (the empirical generalizability with a mini-batch of validation samples). The proposed pseudo supervisor is based on the policy gradient \cite{Sutton_NeurIPS_1999}, which is one type of generic reinforcement learning methods \cite{Baxter_JAIR_2001,Ghavamzadeh_CSDFPS_2003,Silver_ICML_2014}.

Here we formally introduce the concepts related to the proposed pseudo supervisor. An RNFLT map $x_{t}$ at time step $t$ can be viewed as a state $s_{t}\in \mathcal{S}$, while the predicted pseudo label $\tilde{y}_{t}$ can be viewed as an action $a_{t}\in \mathcal{A}$. A reward function $r_{t}: \mathcal{S}\times \mathcal{A} \xrightarrow{} \mathbf{R}$ is defined on the state-action space. The proposed pseudo supervisor that employs a policy to determine pseudo labels is denoted as $\pi_{\omega}: \mathcal{S}\xrightarrow[]{}\mathcal{P}(\mathcal{A})$, where $\mathcal{P}(\mathcal{A})$ are the probability measures on $\mathcal{A}$ and $\omega$ are the parameters of the proposed pseudo supervisor.

In the first place, the classifier takes labeled samples for training. The unlabeled samples will be fed into the pseudo supervisor for predicting pseudo labels. Before updating the weights of the classifier, the empirical generalizability of classifier $f_{\theta_{t}}$ can be quantified with validation samples $(x^{(val)}, y^{(val)})$, that is, $\ell(f_{\theta_{t}}(x^{(val)}), y^{(val)})$ ($\ell_{t}^{(val)}$). Specifically, $\ell$ is the widely-used cross-entropy loss for classification. After updating the weights of the classifier with unlabeled samples and corresponding pseudo labels, given the same mini-batch of validation samples, the empirical generalizability of the updated classifier can be re-quantified again, \ie $\ell(f_{\theta_{t+1}}(x^{(val)}), y^{(val)})$ ($\ell_{t+1}^{(val)}$). Then, the discrepancy between $\ell_{t}^{(val)}$ and $\ell_{t+1}^{(val)}$ is informative to measure the effect of updating the classifier with the unlabeled samples and the pseudo labels. In this sense, it is desired to use in the reward function, \ie
\begin{equation}
    \begin{split}
r_{t} = \max(\exp(\ell_{t}^{(val)} - \ell_{t+1}^{(val)})-1, 0)
\end{split}
\label{eqn:reward}
\end{equation}
In \eqref{eqn:reward}, if $\ell_{t+1}^{(val)}$ is far less than $\ell_{t}^{(val)}$, it implies that the unlabeled samples and pseudo labels lead to a positive effect on the learning process such that the reward increases adaptively.

A trajectory of states, actions, and rewards within a time window $\beta$ will be saved. The expected return of policy optimization can be written as 
\begin{equation}
    \begin{split}
J(\pi_{\omega}) = \mathop{\mathbb{E}_{\pi_{\omega}}} \left( \sum_{k=0}^{} \gamma^{k}r_{k} \right)
\end{split}
\end{equation}
where $\gamma \in [0, 1]$ is the discount rate for the numerical stability of reinforcement learning, and $\beta$ is the time window for updating $\pi_{\omega}$ periodically. According to the policy gradient theorem \cite{Sutton_NeurIPS_1999}, the expectation of the sample gradient is equal to the actual gradient
\begin{equation}
    \begin{split}
\nabla_{\pi_{\omega}}J(\pi_{\omega}) = \mathop{\mathbb{E}_{\pi_{\omega}}} \left[ \sum_{k=0}^{\beta} \gamma^{k}r_{t+k+1}  \nabla_{\pi_{\omega}} \log\pi_{\omega}(a_{t}|s_{t}) \right]
\end{split}
\end{equation}
Then the policy parameterization can be updated according to the gradient ascend rule
\begin{equation}
    \begin{split}
\pi_{\omega} \leftarrow \pi_{\omega} + \eta_{\pi} \nabla_{\pi_{\omega}}J(\pi_{\omega})
\end{split}
\end{equation}
where $\eta_{\pi}$ is the learning rate of $\pi_{\omega}$.





\section{Experiments}
\label{sec:exp}
\subsection{Set-Up \& Implementation Detail}

To optimize our models, we employ the AdamW optimizer \cite{Loshchilov_ICLR_2019} and train all models for 10 epochs in all experiments conducted in this study. For supervised classification models for glaucoma detection, we use a learning rate of 4 $\times~10^{-5}$ and weight decay of 0, and for progression forecasting, we use a learning rate of 2 $\times~10^{-5}$ and weight decay of 0. These hyperparameters are determined to achieve the best baseline performance based on our experiments. The proposed method is trained using a learning rate of 4 $\times~10^{-5}$ and weight decay of 0. The batch sizes used in the supervised model, semi-supervised models, and the proposed method vary due to the different semi-supervised learning contexts \cite{tarvainen2017mean,berthelot2019mixmatch,berthelot2019remixmatch,sohn2020fixmatch,oh2021distribution}. We adopt the batch sizes that result in the best performance for each respective context. To ensure consistency with previous work, we apply data augmentation techniques to the samples in the semi-supervised learning models, including MeanTeacher \cite{tarvainen2017mean}, MixMatch \cite{berthelot2019mixmatch}, ReMixMatch \cite{berthelot2019remixmatch}, FixMatch \cite{sohn2020fixmatch}, and DASO \cite{oh2021distribution}. Specifically, we utilize random crop and resize, as well as random horizontal flip. However, we do not apply these data augmentation techniques to the supervised baseline or the proposed method.

For the glaucoma detection task, we train all models with 7,860 labeled samples and 7,865 unlabeled samples for training. We use  1,047 and 4,287 labeled samples for validation and testing, respectively. For the progression forecasting task, all the models are trained with 2,944 labeled samples and 2,945 unlabeled samples for training. We use 431 and 1,547 labeled samples for validation and testing, respectively. For multi-modality modeling, we simply concatenate the RNFLT images and VF matrices for joint learning using the same architecture as our single-modality model and apply this model to the progression forecasting task. The model is trained and evaluated using the same train-val-test split as our single-modality experiments.

We use the accuracy, F1 score, and AUC on the test sets as our evaluation matrices. 
The code is written in PyTorch \cite{NEURIPS2019_9015} and we use one RTX A6000 GPU for all experiments.

\subsection{Baseline Methods}
Our Supervised baseline is a standard ImageNet pre-trained EffcientNet classifier \cite{Tan_ICML_2021} learning with cross-entropy using only labeled images. For Semi-Supervised baselines, we choose MeanTeacher \cite{tarvainen2017mean}, MixMatch \cite{berthelot2019mixmatch}, ReMixMatch \cite{berthelot2019remixmatch}, FixMatch \cite{sohn2020fixmatch}, and DASO \cite{oh2021distribution} due to their previous powerful performances. MeanTeacher \cite{tarvainen2017mean} is one of the initial methods to enforce models to predict consistent pseudo predictions for labeled and unlabeled data, which construct a target-generating teacher model by averaging the model weights.  
MixMatch \cite{berthelot2019mixmatch}, ReMixMatch \cite{berthelot2019remixmatch}, and FixMatch \cite{sohn2020fixmatch} are consistency-based SSL methods that are based on the data augmentations (i.e., weak and strong augmentation) to make consistent classification predictions between different views of unlabeled data. DASO \cite{oh2021distribution} suggests that those methods fail for imbalanced datasets and proposes a pseudo-labeling framework based on different consistency SSL models by adaptively mixing the linear and semantic pseudo-labels to mitigate the overall bias between majority and minority classes.
We adopt the FixMatch version of DASO, following the default setting reported in the paper \cite{oh2021distribution}. We implement all baseline methods using Pytorch \cite{NEURIPS2019_9015} from DASO's official codes and conducted experiments under the same codebase and experimental protocols for fair comparison. All models are equipped with the same EfficientNetV2-S \cite{Tan_ICML_2021} backbone and their hyper-parameters are specifically tuned to suit our Glaucoma detection and progression forecasting tasks.  

\subsection{Results for Glaucoma Detection}
In Table \ref{tbl:perf_glau}, we evaluate our Pseudo-Sup model against multiple SOTA baselines on the glaucoma detection task. Our model achieves the best performance in terms of accuracy, F1 score, and AUC score. We compute the mean performances over 5 different runs. 
Notably, our model achieves such results without any specific design of image perturbations/augmentations, guaranteeing its simplicity and applicability to different tasks. 
Our method surpasses the previous SOTA method DASO by 6.8\% of accuracy, 11.1\% of F1 score, and 3.7\% of AUC, respectively. The best-performing consistency method FixMatch still worse by 0.4\% of accuracy, 1.2\% of F1 score, and 1.5\% of AUC than our Pseudo-Sup model, indicating the effectiveness of our approach and the challenges of designing appropriate augmentations for RNFLT maps from previous approaches.  We also apply \textbf{weak} augmentations (similar to FixMatch) to our Pseudo-Sup (Pseudo-Sup+Aug) leading to improved performances.  Please note that consistency methods (e.g., DASO, FixMatch, etc.) employ \textbf{both} weak and strong augmentations. 

\subsection{Results for Progression Forecasting}
\textbf{Unimodal Results.} We further evaluate our model on progression forecasting, a more clinically significant task than glaucoma detection, as shown in Table \ref{tbl:perf_glau_1modality}. Accurate forecasting of glaucoma progression plays a vital role in prescribing appropriate treatment plans to efficiently prevent further vision losses and is important for assisting the development of new pharmaceutical medicines to treat glaucoma. Adding weak augmentations further improves the results for both progression tasks. Please note that similar to glaucoma detection experiments, we also compute the mean performances over 5 different runs.

For TD Progression, our model achieves the best 74.47\% of mean AUC, respectively. When compared with consistency-based approaches, our method improves between 0.5\% - 2.1\% in AUC. For MD Fast Progression, our approach consistently achieves the best AUC performance, illustrating the effectiveness and accuracy of our method regardless of any glaucoma progression criteria. 
Given that most of the glaucoma patients are non-progressive, our progression dataset for TD Progression and MD Fast Progression are both highly imbalanced (i.e., most of the samples are labeled as non-progression). Therefore, it is worth noting that our method also outperforms the previous imbalanced SSL SOTA DASO by a large margin in all measures under such imbalanced scenarios. 

\begin{table}
	\centering
	\caption{\label{tbl:perf_glau}
	    Performance on the cross-sectional data for the \textbf{glaucoma detection} task. The baseline method is the supervised model trained with labeled samples only, while pseudo-sup stands for the proposed \textit{pseudo supervisor} method.
	}
 \vspace{-0.3cm}
	\adjustbox{width=.98\columnwidth}{
	\begin{tabular}{@{}cccc@{}}
		\toprule
		\textbf{Model} & \textbf{Acc$\uparrow$} & \textbf{F1$\uparrow$} & \textbf{AUC$\uparrow$} \\
  \hline \hline 
    Baseline & 0.8131 $\pm$ 0.0039 & 0.8057 $\pm$ 0.0033 & 0.8693 $\pm$ 0.0022  \\
		MeanTeacher & 0.8148 $\pm$ 0.0053 & 0.8053 $\pm$ 0.0072 & 0.8541 $\pm$ 0.0040  \\
        MixMatch & 0.6392 $\pm$ 0.0161 & 0.6273 $\pm$ 0.0059 & 0.8451 $\pm$ 0.0062  \\ 
        ReMixMatch & 0.6741 $\pm$ 0.0019 & 0.6716 $\pm$ 0.0018 & 0.8442 $\pm$ 0.0032   \\ 
        DASO & 0.7444 $\pm$ 0.0115 & 0.6929 $\pm$ 0.0050 & 0.8371 $\pm$ 0.0118 \\
        FixMatch & 0.8077 $\pm$ 0.0032 & 0.7912 $\pm$ 0.0021 & 0.8574 $\pm$ 0.0019  \\ \hline
        Pseudo-Sup & \textbf{0.8124 $\pm$ 0.0025} &\textbf{ 0.8038 $\pm$ 0.0046} & \textbf{0.8727 $\pm$ 0.0006} \\
         Pseudo-Sup+Aug & \textbf{0.8224 $\pm$ 0.0101} &\textbf{ 0.8128 $\pm$ 0.0122} & \textbf{0.8818 $\pm$ 0.0020} \\
  \hline \bottomrule	
	\end{tabular}}
\end{table}

\begin{table}
	\centering
	\caption{\label{tbl:perf_glau_1modality}
	    Performance on the longitudinal data with \textbf{single modality} RNFLT for the \textbf{progression forecasting} task. The baseline method is the supervised model trained with labeled samples only, while Pseudo-Sup represents the proposed \textit{pseudo supervisor} method. 
	}
 \vspace{-0.3cm}
	\adjustbox{width=.85\columnwidth}{
	\begin{tabular}{@{}ccc@{}}
		\toprule
    Models  & AUC $\uparrow$ (TD) & AUC $\uparrow$ (MD Fast) \\ \hline \hline
    Baseline & 0.7367 $\pm$ 0.0126 & 0.6677 $\pm$ 0.0017 \\
    MeanTeacher & 0.7367 $\pm$ 0.0126 & 0.6677  $\pm$ 0.0017 \\
    MixMatch & 0.7238 $\pm$ 0.0098 & 0.6518  $\pm$ 0.0069 \\ 
    ReMixMatch & 0.7390  $\pm$ 0.0124  & 0.6595  $\pm$ 0.0083 \\
    DASO & 0.7231  $\pm$ 0.0188 & 0.6787  $\pm$ 0.0148 \\
    FixMatch  & 0.7395  $\pm$0.0152 & 0.6747 $\pm$ 0.0054 \\ \hline
    Pseudo-Sup & 0.7447 $\pm$ 0.0084 & 0.7004 $\pm$ 0.0218 \\
    Pseudo-Sup+Aug & \textbf{0.7477 $\pm$ 0.0074} &\textbf{0.7215 $\pm$ 0.0162}  \\ \hline
  \bottomrule \\
	\end{tabular}}
\end{table}

\textbf{Multimodal Results.}
One of the major contributions of this work is to introduce a multi-modal dataset for glaucoma progression forecasting. To the best of our knowledge, this is the \textit{first} work to explore and benchmark SSL approaches using multimodal model inputs. In Table \ref{tbl:perf_glau_multi_modal}, we show the progression forecasting results using two modalities (RNFLT and VF). Our model achieves the best performance on all evaluation measures. To keep its simplicity and adaptability, we concatenate the RNFLT and the up-scaled VF together as the inputs of our model. Please note that consistency SSL approaches that are based on image augmentations are not included in this table because of the lack of design of augmentation techniques in multimodal medical data. The SOTA results with multimodal glaucoma data enable more accurate and applicable potential computer-aided diagnosis (CAD) systems in real clinical settings.

\subsection{Supervised Benchmarks on Released Data}
In Table \ref{tbl:progression_benchmark}, we show the supervised classification results for the progression forecasting tasks with multiple SOTA supervised CNN and transformer baseline methods, including VGG \cite{simonyan2014very}, ResNet \cite{he2016deep}, ResNext \cite{xie2017aggregated}, WideResNet \cite{zagoruyko2016wide}, EfficientNet \cite{tan2019efficientnet}, ConvNext \cite{liu2022convnet}, ViT \cite{dosovitskiy2020image}, and Swin Transformers \cite{liu2021swin}. This benchmark is conducted on our future data release with 500 patients upon acceptance,  of which 400 patients are used for training and the remaining are used for testing. To the best of our knowledge, this is the first supervised glaucoma progression forecasting benchmark, which aims to provide broader impacts for the computer vision and medical imaging communities to build clinically effective computer-aided diagnosis systems. Moreover, the results are reported with both single RNFLT modality and multi-modality (RNFLT and VF) samples. It is worth noting that the accuracy and F1 are the same on MD fast progression because the testing set only contains 100 samples under highly imbalanced class distributions. 
For more experimental details regarding this supervised progression forecasting benchmark and the supervised results on future release data with 1,000 patients for the glaucoma detection task, please refer to the Supplemental Material.

\subsection{Performances on LAG}
In Table \ref{tbl:benchmark_lag}, we show the performance comparisons on LAG dataset~\cite{li2019attention}. LAG is a large-scale fundus imaging dataset for glaucoma detection. For this experiment, we use 3,854 images for training with half with labels and the remaining without labels, and 1,000 images for testing, with the size of 224 × 224. For fair comparisons, we use the same hyper-parameters and experimental setups as our other benchmarks. 
It is worth noting that our approach surpasses the SOTA methods by a large margin and achieves the highest 98.58\% AUC with only half-labeled data, illustrating the effectiveness and applicability of our proposed Pseudo-Sup in different data and tasks. 

\begin{table}
	\centering
 \caption{Performance comparisons for \textbf{glaucoma detection} task on \textbf{LAG dataset}.}
 \label{tbl:benchmark_lag}
  \vspace{-.15in}
	\adjustbox{width=.95\columnwidth}{
	\begin{tabular}{L{15ex}  C{10ex} C{10ex} C{10ex}}
		\toprule
		\textbf{Model}  & \textbf{Acc$\uparrow$} & \textbf{F1$\uparrow$} & \textbf{AUC$\uparrow$} \\ 
    \hline \hline
        MeanTeacher & 0.8280 & 0.8279 & 0.9082 \\
        MixMatch  & 0.5020 & 0.3378 & 0.9075 \\ 
        ReMixMatch  & 0.6980 &  0.4785 & 0.8369 \\ 
        DASO & 0.7971 & 0.7906 & 0.9164\\
        FixMatch  & 0.8366 & 0.8297 & 0.9198 \\ \hline
       Pseudo-Sup & \textbf{0.9160} & \textbf{0.9158} & \textbf{0.9858}  \\
        \hline
		\bottomrule	
	\end{tabular}}
 
\label{tbl:glau_lag}
\end{table}

\subsection{Ablation Study}

\begin{figure}
  \centering
    \includegraphics[width=0.44\textwidth]{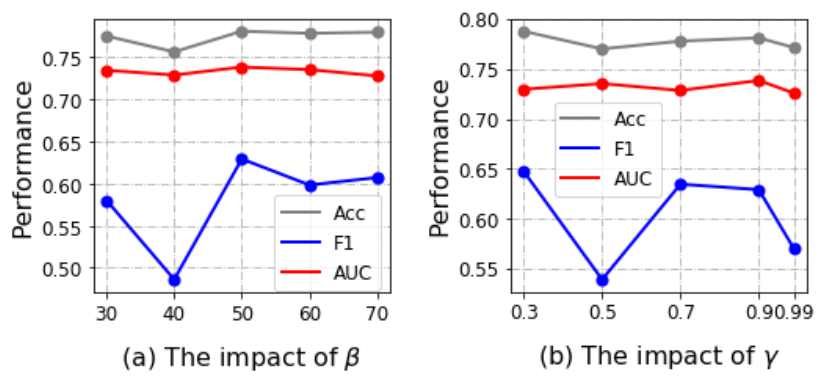}
  \caption{Ablation study of hyperparameters' impact on TD Progression forecasting.} 
  \label{fig:param_ef}
\end{figure}

According to Section \ref{sec:method}, the time window $\beta$ and discount rate $\gamma$ are the most important hyperparameters affecting the proposed pseudo supervisor. The ablation studies of the two hyperparameters are shown in Figure \ref{fig:param_ef}. We can see that the proposed pseudo supervisor behaviors stably with various $\beta$ on AUC. Specifically, when $\beta$ = 50 leads to better performance. The consistent pattern can be also observed in the analysis of $\gamma$. The discount factor is used to achieve a trade-off between the rewards in the distant future and the rewards in the near future. If it is 0, reinforcement learning agents would only look for an optimal policy for the first action. As shown in Figure \ref{fig:param_ef} (b), we find the optimal discount rate at 0.9, which implies that the states, actions, and rewards in the progression task are coherent to a certain degree.

\begin{table}
	\centering
	\caption{\label{tbl:perf_glau_multi_modal}
	    Performance on the longitudinal data with \textbf{two modalities} (RNFLT and VF) for the \textbf{progression forecasting} task. The baseline method is the supervised model trained with labeled samples only, while pseudo-sup stands for the proposed \textit{pseudo supervisor} method. Please note that most methods in Table \ref{tbl:perf_glau_1modality} are based on weak/strong augmentations \cite{sohn2020fixmatch,berthelot2019mixmatch,berthelot2019remixmatch} requiring RGB channels, which are not adaptable for multi-modal learning.
	}
 \vspace{-0.3cm}
	\adjustbox{width=.8\columnwidth}{
		\begin{tabular}{@{}ccc@{}}
		\toprule
    Models  & AUC (TD) & AUC (MD Fast) \\ \hline \hline
    Baseline & 0.7343 $\pm$ 0.0098 & 0.6814 $\pm$ 0.0081 \\
    MeanTeacher   & 0.7436 $\pm$ 0.0133 & 0.6697 $\pm$ 0.0035 \\
Pseudo-Sup & \textbf{0.7517 $\pm$ 0.0041} & \textbf{0.7265 $\pm$ 0.0095}  \\ \hline
  \bottomrule 
	\end{tabular}}
\end{table}

\begin{table}
	\centering
	\caption{\label{tbl:progression_benchmark}
	    Performance of different \textbf{supervised methods} on the longitudinal data with single/multi-modality for the \textbf{progression forecasting tasks} on the released 500 glaucoma progression data.
	}
 \vspace{-0.3cm}
	\adjustbox{width=1\columnwidth}{
	\begin{tabular}{L{16ex} C{10ex} C{10ex} C{10ex} C{10ex} C{10ex} C{10ex}}
		\toprule
             & \multicolumn{3}{c}{TD Progression} & \multicolumn{3}{c}{MD Fast Progression} \\
            \cmidrule(lr){2-4} \cmidrule(lr){5-7}
		\textbf{Model} & \textbf{Acc$\uparrow$} & \textbf{F1$\uparrow$} & \textbf{AUC$\uparrow$} & \textbf{Acc$\uparrow$} & \textbf{F1$\uparrow$} & \textbf{AUC$\uparrow$} \\
		\cmidrule(lr){1-1} \cmidrule(lr){2-2} \cmidrule(lr){3-3} \cmidrule(lr){4-4} \cmidrule(lr){5-5} \cmidrule(lr){6-6} \cmidrule(lr){7-7}  
        
        
        VGG \cite{simonyan2014very} & 0.78/0.80 & 0.75/0.74 & 0.84/0.79 & 0.98/0.98 & 0.49/0.49 & 0.78/0.68   \\ 
        ResNet \cite{he2016deep} & 0.75/0.74 & 0.64/0.67 & 0.74/0.75 & 0.98/0.98 & 0.49/0.49 & 0.71/0.59\\
        ResNext \cite{xie2017aggregated} & 0.78/0.77 & 0.65/0.68 & 0.78/0.76 & 0.98/0.98 & 0.49/0.49 & 0.61/0.80\\
        WideResNet \cite{zagoruyko2016wide} & 0.77/0.79 & 0.71/0.75 & 0.77/0.80 & 0.98/0.98 & 0.49/0.49 & 0.71/0.64\\ 
        EfficientNet \cite{tan2019efficientnet} & 0.73/0.78 & 0.66/0.73 & 0.76/0.79 & 0.98/0.98 & 0.49/0.49 & 0.55/0.66\\ 
        ConvNext \cite{liu2022convnet} & 0.74/0.81 & 0.62/0.77 & 0.78/0.81  & 0.98/0.98 & 0.49/0.49 & 0.84/0.62\\ 
        ViT \cite{dosovitskiy2020image} & 0.73/0.77 & 0.54/0.71 & 0.68/0.79 & 0.98/0.98 & 0.49/0.49 & 0.92/0.69 \\
        Swin \cite{liu2021swin} & 0.71/0.77 & 0.49/0.72 & 0.68/0.77 & 0.98/0.98 & 0.49/0.49 & 0.87/0.57 \\
   
		\bottomrule	
	\end{tabular}}
\end{table}

\subsection{Experiment on Racial Groups}

 The patients in our dataset are from three racial groups (White (0.88 AUC), Black (0.83 AUC), and Asian (0.85 AUC)). In our experiments, we observe that black patients have the lowest AUC yet the highest glaucoma prevalence~\cite{rudnicka2006variations, friedman2006prevalence}. Therefore, our dataset is well-suited for future fairness studies.
\section{Conclusion}
In this paper, we have made two contributions. First, we have developed a generalization-reinforced semi-supervised learning model termed pseudo supervisor for glaucoma detection and progression forecasting. Specifically, the pseudo supervisor predicts pseudo labels following a policy that optimizes the generalization error in training the classifier. Our pseudo supervisor model overall demonstrates superior performance over a number of SOTA methods. Second, we release a multimodal and multi-task dataset for glaucoma detection and progression forecasting with SOTA 3D OCT imaging data. Our dataset is the \textit{largest} by far for glaucoma detection with 3D OCT data and the \textit{first} for progression forecasting task. More importantly, our dataset entails demographic information including gender and race, which is generally unavailable in existing public glaucoma datasets. Given the diverse racial representation in our dataset as detailed in Section \ref{data_section}, our dataset is well equipped for potential fairness learning studies, especially considering the significant glaucoma prevalence disparity between Black and other races \cite{rudnicka2006variations, friedman2006prevalence}. 

\section*{Acknowledgment}

This work was supported by NIH R00 EY028631, Research to Prevent Blindness International Research Collaborators Award, Alcon Young Investigator Grant, NIH R01 EY030575, and NIH P30 EY003790.

	%
	%

{\small
\bibliographystyle{ieee_fullname}
\bibliography{egbib}
}


\clearpage
\newpage
\begin{center}
\textbf{\Large Appendix}
\end{center}

\section{Implementation Detail for Supervised Benchmarks}

For the optimization, we use AdamW optimizer \cite{Loshchilov_ICLR_2019} and train all the supervised models with 20 epochs throughout all the experiments. We use learning rate 2e-5 and weight decay 1e-5 with a batch size of 12 for all the supervised classification models for all methods in supervised progression forecasting and glaucoma detection benchmarks. All supervised classification models are trained using BCE loss. 
For ViT, we used their ViT-B-16 architecture. For EfficientNet, we use their EfficientNetV2-S architecture. For Swin transformer, we use their Swin-base architecture. For ResNet, we use their ResNet50 architecture. For VGG, we use their VGG-11 architecture. For ResNeXt, we use their ResNeXt-101 64$\times$4d architecture. For WiderResNet, we use their Wide ResNet-50-2 architecture. For ConvNeXt, we use their ConvNeXt Tiny architecture. We initialize all models with pre-trained imagenet weights. 
All code is written in PyTorch \cite{NEURIPS2019_9015} and we use one RTX A6000 GPU for all experiments.

\begin{table}[!t]
	\centering
	\caption{\label{tbl:glaucoma_benchmark}
	    Performance of different \textit{supervised methods} on the cross-sectional data with single modality RNFLT for the \textit{glaucoma detection} task on the released \textbf{1,000} glaucoma detection data.
	}
 \vspace{-0.3cm}
	\adjustbox{width=0.85\columnwidth}{
	\begin{tabular}{L{20ex} L{10ex} L{10ex} L{10ex}}
		\toprule
		\textbf{Model} & \textbf{Acc$\uparrow$} & \textbf{F1$\uparrow$} & \textbf{AUC$\uparrow$}  \\
	\toprule
        
        
        VGG \cite{simonyan2014very} & 0.80 & 0.79 & 0.86  \\ 
        ResNet \cite{he2016deep} & 0.84 & 0.83 & 0.87 \\
        ResNext \cite{xie2017aggregated} & 0.82 & 0.81 & 0.89\\
        WideResNet \cite{zagoruyko2016wide} & 0.83 & 0.84 & 0.89 \\ 
        EfficientNet\cite{tan2019efficientnet} & 0.85 & 0.85  & 0.90 \\ 
        ConvNext \cite{liu2022convnet} & 0.80 & 0.79 & 0.86 \\ 
        ViT \cite{dosovitskiy2020image} & 0.65 & 0.67 & 0.75  \\
        Swin \cite{liu2021swin} & 0.74 & 0.73 & 0.78  \\
   
		\bottomrule	
	\end{tabular}}
\end{table}

\section{Supervised Benchmarks on Released Data}
In Table\ref{tbl:glaucoma_benchmark}, we show the supervised classification results for the \textbf{glaucoma detection} task with multiple SOTA supervised CNN and transformer baseline methods, including VGG \cite{simonyan2014very}, ResNet \cite{he2016deep}, ResNext \cite{xie2017aggregated}, WideResNet \cite{zagoruyko2016wide}, EfficientNet \cite{tan2019efficientnet}, ConvNext \cite{liu2022convnet}, ViT \cite{dosovitskiy2020image}, and Swin Transformers \cite{liu2021swin}. This \textbf{cross-sectional} benchmark is conducted on our future cross-sectional data release with 1,000 patients upon acceptance, of which 800 patients are used for training and the remaining 200 are used for testing. 
To the best of our knowledge, this is the largest supervised glaucoma detection benchmark with 3D OCT imaging data (i.e., RNFLT). Such large-scale public 3D OCT dataset will encourage researchers to build clinically effective (3D OCT source data) and efficient (post-processed 2D RNFLT map from the 3D data) glaucoma CAD systems. As shown in the table, transformed-based architectures tend to obtain worse performance than CNN-based architectures, and we articulate this due to that transformed-based architectures are often data-hungry and require a relatively larger amount of training data. EfficientNet is the best-performing method, followed by WideResNet and ResNext.  

\subsection{Data Density Distribution}

\begin{figure}[!htbp]
  \centering
    \includegraphics[width=0.4\textwidth]{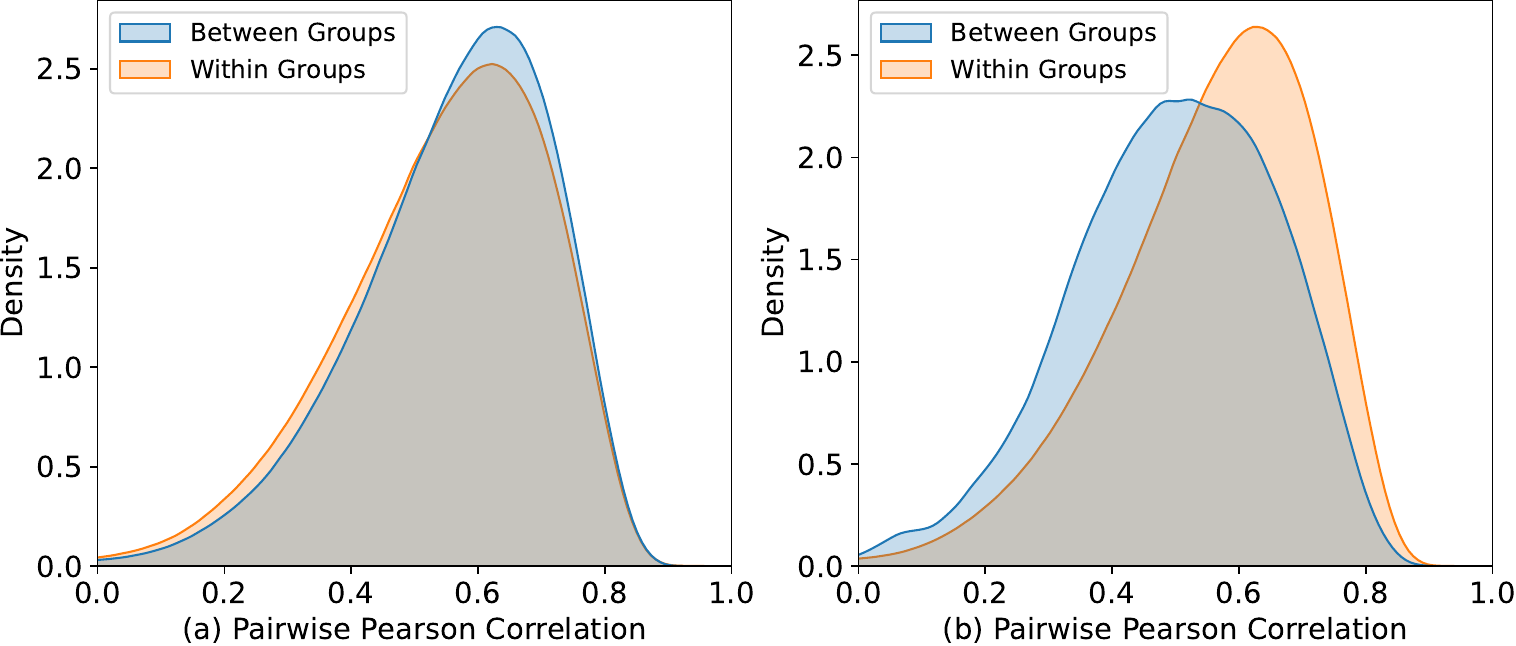}
     \vspace{-0.3cm}
     \caption{The pointwise similarity between RNFLT maps within the same label groups versus the pointwise similarity between RNFLT maps between different label groups.} 
  \label{vardist}
   \vspace{-0.5cm}
\end{figure}

As shown in Fig. \ref{vardist}, the density distribution of correlations between RNFL thickness (RNFLT) maps within groups of glaucoma and non-glaucoma is largely overlapped with the one between RNFLT maps between glaucoma and non-glaucoma groups. The same is observed for progression versus non-progression.

\end{document}